\documentclass[10pt,twocolumn,letterpaper]{article}

\usepackage[pagenumbers]{iccv} 
\usepackage{booktabs}
\usepackage{tabularx}
\usepackage{multirow}
\usepackage{amsmath}
\usepackage{algorithm}
\usepackage{xcolor}
\usepackage{algpseudocode}
\usepackage{graphicx}

\definecolor{iccvblue}{rgb}{0.21,0.49,0.74}
\usepackage[pagebackref,breaklinks,colorlinks,allcolors=iccvblue]{hyperref}

\title{Personalized Federated Learning via Learning Dynamic Graphs}

\author{
    Ziran Zhou\textsuperscript{1} \quad Guanyu Gao\textsuperscript{1}\thanks{Corresponding Author} \quad Xiaohu Wu\textsuperscript{2} \quad Yan Lyu\textsuperscript{3} \\
    \texttt{\small \{zzrac@njust.edu.cn, gygao@njust.edu.cn, xiaohu.wu@bupt.edu.cn, lvyanly@seu.edu.cn\}} \\
    \textsuperscript{1}Nanjing University of Science and Technology \\ \textsuperscript{2}Beijing University of Posts and Telecommunications\\ 
    \textsuperscript{3}Southeast University
}

\begin{document}
\maketitle
\begin{abstract}
Personalized Federated Learning (PFL) aims to train a personalized model for each client that is tailored to its local data distribution, effectively addressing the issue where the global model in federated learning fails to perform well on individual clients due to variations in their local data distributions. Most existing PFL methods focus on personalizing the aggregated global model for each client, neglecting the fundamental aspect of federated learning: the regulation of how client models are aggregated. Additionally, almost all of them overlook the graph structure formed by clients in federated learning. In this paper, we propose a novel method, Personalized Federated Learning with Graph Attention Network (\texttt{pFedGAT}), which captures the latent graph structure between clients and dynamically determines the importance of other clients for each client, enabling fine-grained control over the aggregation process.  We evaluate \texttt{pFedGAT} across multiple data distribution scenarios, comparing it with twelve state-of-the-art methods on three datasets—Fashion MNIST, CIFAR-10, and CIFAR-100—and find that it consistently performs well.
\end{abstract}    
\section{Introduction}
\label{sec:intro}

In today’s digital age, the rapid growth and distributed nature of data present significant challenges for training machine learning models. Federated Learning, an emerging collaborative learning framework, seeks to leverage data distributed across multiple clients to train a global model without requiring raw data sharing, thus preserving data privacy. Since it was proposed in 2017 \cite{fedavg}, Federated Learning has garnered significant attention and found applications in a variety of domains, including computer vision \cite{fedvision, FedCV}, natural language processing \cite{Fednlp, Fedadapter}, and healthcare \cite{swarm}. However, real-world data often exhibits non-independent and identically distributed (Non-IID) characteristics, which means that there are substantial discrepancies in the data distribution across clients. This creates challenges for traditional federated learning, as it struggles to achieve optimal performance under such conditions.

To address this issue, Personalized Federated Learning (PFL) has emerged \cite{mocha, perfedavg}, with the primary goal of tailoring models to better align with the local data distribution of each client, thereby enhancing performance in Non-IID environments. For example, \texttt{Per-FedAvg}\cite{perfedavg} uses meta-learning on local clients to enable the aggregated global model to quickly adapt to the local data distribution. \texttt{MOCHA}\cite{mocha} employs multi-task learning to train a personalized model for each client. \texttt{VIRTUAL}\cite{VIRTUAL} further utilizes a Bayesian approach to handle non-convex models. \texttt{pFedBayes}~\cite{pFedBayes} trains personalized Bayesian models at each client tailored to the unique complexities of the clients' datasets and efficiently collaborating across these clients. \texttt{APFL}\cite{APFL} applies an adaptive algorithm to learn a personalized model for each client, which is a mixture of the optimal local model and the global model. \texttt{pFedHN}\cite{pfedhn} uses a hypernetwork to generate personalized models for each client.

However, the above works focus on training personalized models that adapt to local data distributions, 
 neglecting the basic element of federated learning: the aggregation between client models. \texttt{FedPer}~\cite{fedper} retains the personalized layers of each client and only aggregates the base layers on the server. \texttt{FedAH}~\cite{FEDAH} addresses the loss of information in traditional personalized head approaches by performing element-wise aggregation of local and global model heads, thereby enhancing overall model performance.  \texttt{FedAMP}~\cite{FedAMP} uses an attention mechanism to increase the collaboration intensity between clients with similar data distributions, while \texttt{HeurFedAMP}~\cite{FedAMP} further allocates model parameters entirely based on the similarity of client model parameters. \texttt{CFL}~\cite{CFL} forms small clusters of clients with high task similarity and conducts collaboration within the clusters. \texttt{IFCA}~\cite{IFCA} builds K clusters, assigns each client to one of the K clusters, and conducts collaboration within the clusters. \texttt{FedGroup}~\cite{Fedgroup} uses the Kmeans++ algorithm to cluster based on local client model updates. \texttt{FedEgoists}~\cite{FedEgoists} optimizes collaboration in the federated learning process by considering data complementarity and competitive relationships among participants.  Although these works have improved the aggregation methods between client models, they only make coarse-grained improvements and do not fine-grained regulate the aggregation methods between client models.

In order to enable fine-grained adjustment of the aggregation methods between client models, and utilize the graph structural information between client models in federated learning, we propose the \texttt{pFedGAT} algorithm. We deploy a Graph Attention Network  
 (GAT)~\cite{GAT} model on the server side, which captures the graph structural information between them based on the model parameters uploaded by each client, and dynamically calculates the importance of other clients to the current client. For a client, it will collaborate as much as possible with the clients that are more important to itself.
In summary, our contributions are:




\begin{itemize}
\item We propose an innovative personalized federated learning approach, \texttt{pFedGAT}, which uses GAT to dynamically adjust the collaboration relationships between clients based on the graph structure formed by their models. To the best of our knowledge, we are the first to employ GAT to refine model aggregation strategies.

\item We develop a novel end-to-end training paradigm that optimizes GAT-based aggregation using client-side loss feedback, ensuring robust and adaptive model personalization across diverse data distributions.

\item We evaluate \texttt{pFedGAT} across three datasets under varying data distribution scenarios. Experimental results demonstrate that our methods consistently outperform state-of-the-art approaches in all tested conditions.
\end{itemize}

\section{Related Work}
\label{sec:formatting}

\subsection{Federated Learning}
Federated Learning  is an emerging machine learning paradigm that enables the collaborative training of models across multiple clients, each with isolated datasets, while preserving privacy. The goal is to train a global model that performs well across all clients. \texttt{FedAvg} \cite{fedavg}, one of the most widely used FL algorithms, aggregates local models from clients based on the size of their respective datasets to obtain a global model. However, in real-world scenarios, the data distributions among clients often follow a Non-IID pattern, which makes simple aggregation methods insufficient for addressing the unique needs of each client. To overcome this challenge, several approaches \cite{FedProx, scaffold, MooN, pFedMe, Ditto,FedNova, LG-FedAvg} have been proposed.

For instance, \texttt{FedProx} \cite{FedProx} introduces a regularization term that penalizes deviations between local and global model parameters, thereby preventing local updates from straying too far from the global model. \texttt{SCAFFOLD} \cite{scaffold} uses control variates to mitigate the biases in model updates caused by Non-IID data, improving both accuracy and convergence speed. \texttt{MOON} \cite{MooN} leverages contrastive learning to minimize the representational gap between local and global models, reducing the risk of local models diverging excessively. \texttt{LG-FedAvg} \cite{LG-FedAvg} proposes a hybrid approach that combines local and global representations, enhancing both personalization and generalization by learning compact local representations and optimizing the global model jointly. \texttt{FedNova} \cite{FedNova} adjusts aggregation weights based on both data volume and local update frequency for each client.

However, despite these advancements,  they still fall short in addressing the need for personalized models tailored to clients with highly heterogeneous data distributions.

\subsection{Personalized Federated Learning}


 Personalized federated learning  aims to deliver tailored machine learning models for individual users, adapting to their unique data distributions and personalized needs while preserving privacy. Existing methods address this challenge through diverse strategies. For instance, \texttt{FedPer}~\cite{fedper} separates models into shared and personalized layers, striking a balance between global knowledge sharing and local data adaptation. Similarly, approaches employing knowledge distillation, such as \texttt{FedMD}~\cite{FedMd}, \texttt{FedDF}~\cite{FedDF}, \texttt{FedGKT}~\cite{FedGKT} and \texttt{FedGMKD}~\cite{FedGMKD}, enable knowledge transfer across clients: \texttt{FedMD} fine-tunes models using local data after aligning with a global model via a public dataset, \texttt{FedDF} adjusts distillation weights dynamically to fit client-specific requirements, \texttt{FedGKT} groups clients to refine models based on group traits and local data and \texttt{FedGMKD} combines knowledge distillation and differential aggregation for efficient
prototype-based personalized FL.

Other methods focus on leveraging client similarity or structural adaptations. \texttt{FedAMP}~\cite{FedAMP} dynamically adjusts aggregation weights based on inter-client similarity, while \texttt{CFL}~\cite{CFL} clusters clients by data distribution similarity, conducting federated learning within each cluster. \texttt{pFedHN}~\cite{pfedhn} trains a server-side hypernetwork to generate personalized model parameters for each client, and \texttt{FedROD}~\cite{FedRoD} decouples models into a shared global representation layer and client-specific heads. Additionally, \texttt{Per-FedAvg}~\cite{perfedavg} integrates meta-learning to allow rapid personalization with few local updates, \texttt{pFedMe}~\cite{pFedMe} enforces an $\ell_{2}$ distance constraint between personalized and global models, and \texttt{KNN-PER}~\cite{KNN-PER} uses K-nearest neighbors to identify client similarities for targeted fine-tuning.

Despite these advances, the aforementioned methods overlook the graph-based relationships among clients in federated learning. In  \texttt{pFedGAT}, we employ Graph Attention Network (GAT)~\cite{GAT} to capture inter-client dependencies, enabling each client to collaborate more effectively with similar peers rather than relying on simplistic aggregation.

\subsection{Graph Learning}
Graph Neural Network (GNN)~\cite{GNN} is a class of neural network model designed specifically for processing graph-structured data. Graph Convolutional Network (GCN)~\cite{GCN} and Graph Attention Network (GAT)~\cite{GAT} are likely the most widely used variants of GNN. GCN adapts the concept of convolution to graphs, achieving simple and efficient neighbor aggregation via a normalized adjacency matrix. However, GCN treats all neighbors of a node equally, failing to differentiate their varying importance. GAT addresses this by incorporating an attention mechanism, enabling the model to dynamically learn the significance of neighboring nodes. 

\texttt{SFL}~\cite{sfl} is the first federated learning algorithm to consider graph relationships among clients. However, it employs GCN to capture these inter-client relationships, which struggles to dynamically extract their varying significance. In contrast, \texttt{pFedGraph}\cite{pFedGraph} and \texttt{FedAGHN}~\cite{fedaghn} leverage quadratic programming and hypernetworks, respectively, to model collaborative relationships among clients, yet they overlook the underlying graph structure between clients.

In our proposed \texttt{pFedGAT} and \texttt{pFedWGAT}, we represent each client as a node in a graph (see Section \ref{sec:GraphConstruction} for details on the graph structure) and employ GAT to dynamically determine the importance of other clients to a given client, thereby facilitating enhanced collaboration among clients in federated learning.
\begin{figure}
    \centering
    \includegraphics[width=\linewidth]{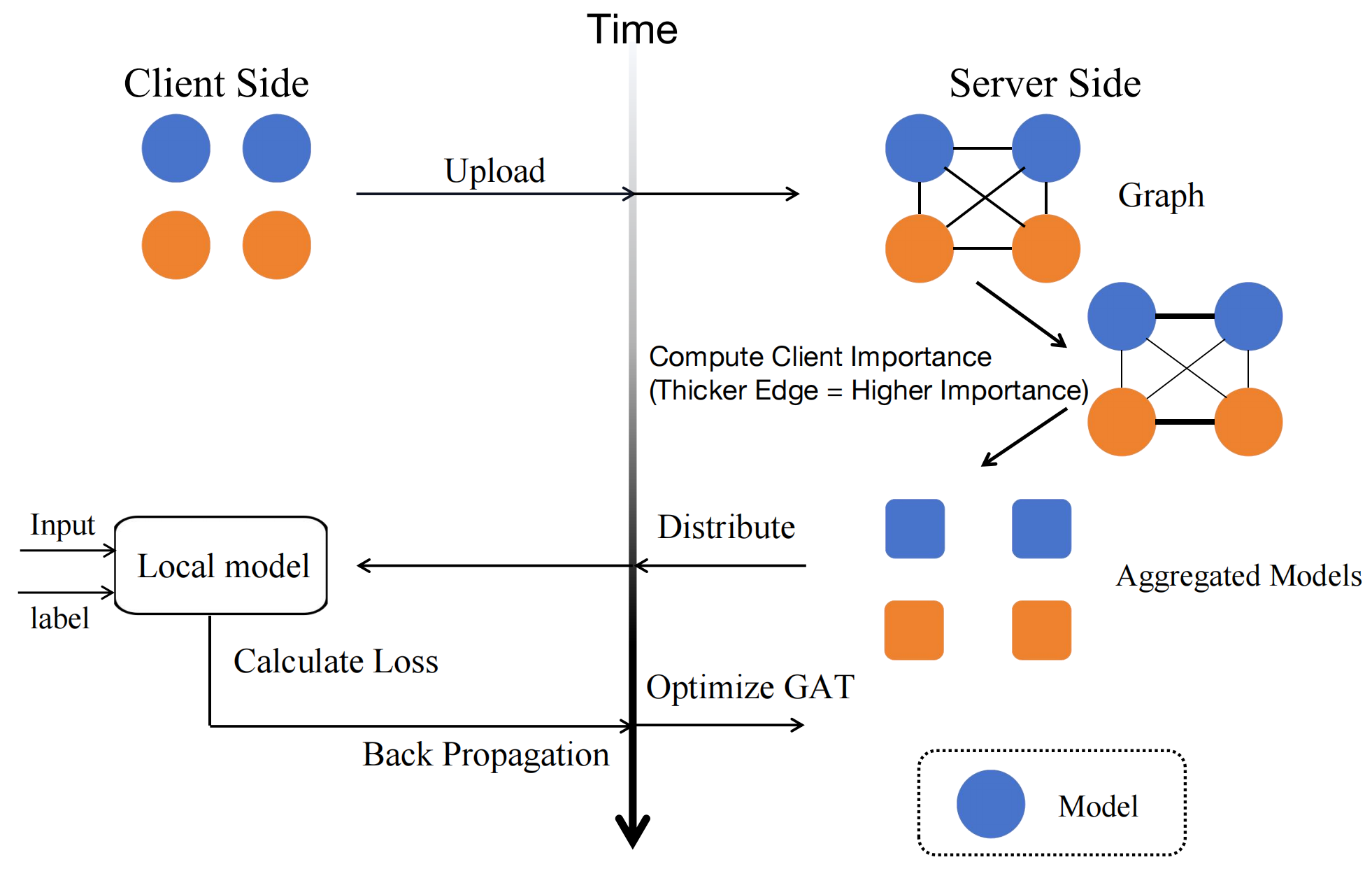}
    \caption{Workflow of pFedGAT}
    \label{fig:workflow}
\end{figure}

\vspace{0.1cm}
\section{Method}

In this section, we begin by formulating the Personalized Federated Learning (PFL) problem, then introduce our \textit{Personalized Federated Learning with Graph Attention Neural Network} (\texttt{pFedGAT}) algorithm. Finally, we describe our novel end-to-end optimization strategy.

\subsection{Problem Formulation}
Traditional federated learning aims to train a global model across a federation of clients with similar data distributions. For instance, in the widely used \texttt{FedAvg} ~\cite{fedavg} algorithm, the training objective of \textit{N} clients is defined as:

\begin{equation}
   \mathop{min}\limits_{\theta} \space \mathcal{L}(\theta) = \sum_{i=1}^{N}p_{i}\mathcal{L}_{i}(\theta), 
\end{equation}
where $\theta$ denotes the global model, $\mathcal{L}_{i}(\cdot)$ represents the local objective of client $i$, and the weight $p_{i}$ is typically set as $p_{i} = n_{i}/n$ with $n=\sum_{k}n_{k}$, where $n_{k}$ is the number of data samples on client $k$. 

However, PFL shifts the focus to training a unique model $\theta_{i}$ for each client $i$ to optimize performance on their specific local data distributions $\mathcal{D}_{i}$. The training objective of PFL is formulated as

\begin{equation}
    \mathop{min}\limits_{\theta_{1},\theta_{2},\cdots,\theta_{N}} \sum_{i=1}^{N}\mathcal{L}_{i}(\theta_{i};\mathcal{D}_{i}),
\end{equation}
where $N$ is the total number of clients in FL. A straightforward approach woulde be to train each client's model using only their local data to minimize $\mathcal{L}_{i}(\theta_{i};\mathcal{D}_{i}^{train})$, where $\mathcal{D}_{i}^{train}$ is the training set of the local data distribution of client $i$. However, given the limited data available on each client, this approach is highly prone to overfitting on their local data. Consequently, collaboration among clients during training is crucial to enhance generalization. 

A key challenge in PFL is determining an effective collaboration strategy. In IID scenariors, where clients have similar data distributions ($\mathcal{D}_{i}$=$\mathcal{D}_{j}$, i$\ne$j), \texttt{FedAvg} can be directly to aggregate models. However, in non-IID scenariors, where data distributions vary across clients ($\mathcal{D}_{i}$$\ne$$\mathcal{D}_{j}$, i$\ne$j), naive collaboration may degrade local model performance. 

To address this challenge in PFL, we must dynamically adjust the model aggregation weights among clients to mitigate the impact of data heterogeneity. Specifically, clients should prioritize collaboration with those having similar data distributions to gain more beneficial knowledge, while minimizing interactions with clients whose data distributions significantly differ, thereby reducing the assimilation of detrimental knowledge. Additionally, it is crucial to prevent clients from overfitting to their local data, achieving a balance between personalization and generalization capability. In the subsequent sections, we will elaborate on our proposed solution in detail.

\subsection{pFedGAT Algorithm}

Since accurately delineating find-grained cooperative relationships among clients is fundamental to PFL in non-IID scenariors, we propose a novel aggregation-based PFL method, named Personalized Federated Learning with Graph Attention Neural Network Network (\texttt{pFedGAT}). In this approach, we deploy a GAT~\cite{GAT} model on the server to effectively capture inter-client relationships, enabling a more precise allocation of interaction weights between clients.

\subsubsection{Graph Construction}
\label{sec:GraphConstruction}

Consider a federated learning system with \(N\) clients, denoted as \(C_1, C_2, \ldots, C_N\), each uploading their model parameters \(\theta_1^t, \theta_2^t, \ldots, \theta_N^t\) to the server at the \(t\)-th communication round. To construct node features, these parameters are first flattened into vectors and normalized using Layer Normalization (LayerNorm) to ensure numerical stability and consistency across clients. The resulting feature vector for client \(i\) is defined as:
\begin{equation}
    h_i^t = \text{LayerNorm}(\text{Flatten}(\theta_i^t)), \quad h_i^t \in \mathbb{R}^{d}, \label{eq:flatten}
\end{equation}
where \(d\) represents the dimensionality of the flattened model parameters.

These clients are then modeled as nodes in a graph, with their feature vectors \(h_i^t\) serving as node features. In a traditional Graph Attention Network (GAT), the adjacency matrix typically uses binary values (0 or 1) to indicate the presence or absence of edges between nodes. In contrast, our \texttt{pFedGAT} introduces a distinct topological structure: the initial adjacency matrix \(A\) is fully connected, with all entries set to 1. This design assumes equal importance among all clients at the outset, enabling the GAT to dynamically learn client relationships without predefined constraints, setting the stage for personalized aggregation in subsequent steps.

\subsubsection{Attention-Based Weight Assignment}
\label{sec:WeightAssignment}

To perform personalized parameter aggregation, we first project the node features \(h_i^t \in \mathbb{R}^{d}\) into a higher-level space using a shared weight matrix \(W \in \mathbb{R}^{d' \times d}\):
\begin{equation}
    z_i^t = W h_i^t, \quad \forall i \in \{1, \ldots, N\},
\end{equation}
where \(d'\) is the output dimension, enhancing the representational capacity of the features.

Next, for each pair of nodes \((i, j)\), we compute raw attention scores to capture their relationships:
\begin{equation}
    e_{ij}^t = \text{LeakyReLU}(a^T [z_i^t || z_j^t]), \label{eq:attention_scores}
\end{equation}
where \(\text{LeakyReLU}\) is the activation function, \(a \in \mathbb{R}^{2d'}\) is a learnable attention parameter, and \(||\) denotes vector concatenation.

The refined attention scores are then normalized using a softmax operation:
\begin{equation}
    a_{ij}^t = \frac{\exp(e_{ij}^t)}{\sum_{k=1}^{N} \exp(e_{ik}^t)}, \quad \forall j \in \{1, 2, \ldots, N\}, \label{eq:attention}
\end{equation}
yielding the attention coefficients \(a_{ij}^t\).

Following the traditional GAT approach, we employ \(K\) independent attention heads, each computing an attention matrix \(a^{t,k}\) (\(k \in [1, K]\)) using Equation \ref{eq:attention}. Unlike traditional GAT, which typically includes an output layer for downstream tasks, our method focuses solely on parameter allocation. Thus, we define the allocation matrix \(R \in \mathbb{R}^{N \times N}\) as the average across all attention heads:
\begin{equation}
    R_{ij}^t = \frac{1}{K} \sum_{k=1}^{K} a_{ij}^{t,k}, \label{eq:allocation_matrix}
\end{equation}

For each client \(i\), the personalized model parameters for the next round are aggregated as:
\begin{equation}
    \theta_i^{t+1} = \sum_{j=1}^{N} R_{ij}^t \cdot \theta_j^t, \label{eq:aggregation}
\end{equation}

Through this process, we generate tailored model parameters for each client, effectively achieving personalization in federated learning.





\subsection{End-to-End Optimization with Client Feedback}
\label{sec:EndToEndOptimization}

In the previous section, we described how the server-side GAT generates personalized model parameters for each client. Here, we introduce an end-to-end optimization approach to refine the GAT, addressing the limitations of static aggregation under dynamic client distributions.

After generating personalized parameters \(\theta_i^{t+1}\) using Equation \ref{eq:aggregation}, these are distributed to their respective clients. Each client evaluates \(\theta_i^{t+1}\) on its local test set \(\mathcal{D}_i^{\text{test}}\), a held-out subset of its data, and computes the test loss \(\mathcal{L}_i\). Unlike traditional methods, this loss is not used to update \(\theta_i\) locally but is instead uploaded to the server to optimize the GAT. The total loss is aggregated as:
\begin{equation}
    \mathcal{L} = \sum_{i=1}^{N} \mathcal{L}_i,
    \label{eq:loss_all}
\end{equation}

The server then updates the GAT’s learnable parameters \(W\) and \(a\) via a single gradient descent step:
\begin{equation}
    \begin{cases}
        W \leftarrow W - \eta \frac{\partial \mathcal{L}}{\partial W}, \\
        a \leftarrow a - \eta \frac{\partial \mathcal{L}}{\partial a},
    \end{cases}
    \label{eq:optimize}
\end{equation}
where \(\eta\) is the learning rate, and gradients are derived through backpropagation from \(\mathcal{L}\). Since each client uploads only a scalar \(\mathcal{L}_i\) (e.g., 4 bytes), the additional communication overhead is negligible compared to model parameters. This optimization enables the GAT to adapt its aggregation strategy based on real-time client feedback, enhancing personalization across diverse data distributions.The complete algorithm workflow can be found in Algorithm 
\ref{alg:pfedgat}.

\begin{algorithm}
\caption{\texttt{pFedGAT}  Algorithm}
\label{alg:pfedgat}
\begin{algorithmic}[1]
\Require Number of communication rounds \(T\), \(N\) clients with local datasets \(\mathcal{D}_i\),  initial local models \(\theta_i^0\), local training epochs \(E\),  local learning rate \(\eta_1\), GAT learning rate \(\eta_2\)

\For{\(t = 0\)  to  \(T-1\)} 
    \State \textcolor{blue}{\texttt{/*\ Local Training*/}}
    \For{each client \(i = 1\) to \(N\) \textbf{in parallel}} 
        \For{epoch = 1 to \(E\)} 
            \State \(\theta_i^t \gets \theta_i^t - \eta_1 \nabla_{\theta_i^t} \mathcal{L}_i(\theta_i^t; \mathcal{D}_i)\) 
        \EndFor
        \State Client \(i\) uploads \(\theta_i^t\) to server 
    \EndFor
    \State \textcolor{blue}{\texttt{/*\ Model Allocation*/}}
    \For{each attention head} 
        \For{each client pair \((i, j)\)} 
            \State Compute attention score \(e_{ij}\) using Eq.~(\ref{eq:attention_scores}) 
        \EndFor
    \EndFor
    \State Compute allocation matrix \(R\) using Eq.~(\ref{eq:allocation_matrix})
    \State Server aggregates models using Eq.~(\ref{eq:aggregation}) and distributes them to each client
    \State \textcolor{blue}{\texttt{/*\ Optimize GAT*/}}
    \For{each client \(i = 1\) to \(N\) \textbf{in parallel}}
        \State Compute \(\mathcal{L}_i\) and upload it to server
    \EndFor

    \State Server computes total \(\mathcal{L}\) using Eq.~(\ref{eq:loss_all})
    \State Update \(W\) and \(a\) using Eq.~(\ref{eq:optimize})
\EndFor
\end{algorithmic}
\end{algorithm}

\vspace{0.1cm}
\section{Experiments}
\label{ref:experiments}

\subsection{Experiments Setup}

\begin{figure}[t] 
    \centering
    \begin{subfigure}[b]{0.236\textwidth} 
        \centering
        \includegraphics[width=\linewidth, height=4cm]{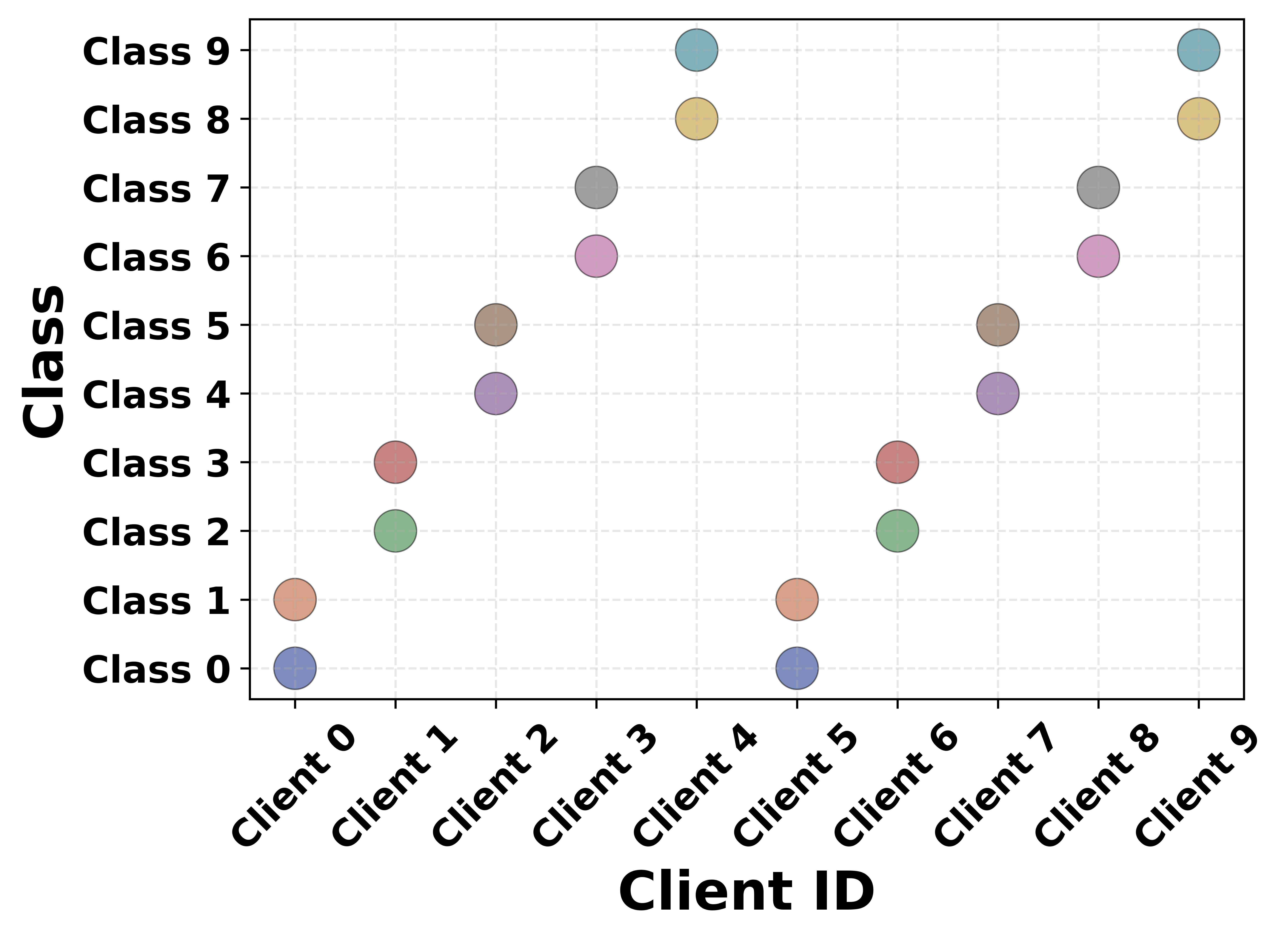}
        \caption{Pathological}
        \label{fig:image1}
    \end{subfigure}
    \hfill
    \begin{subfigure}[b]{0.236\textwidth}
        \centering
        \includegraphics[width=\linewidth, height=4cm]{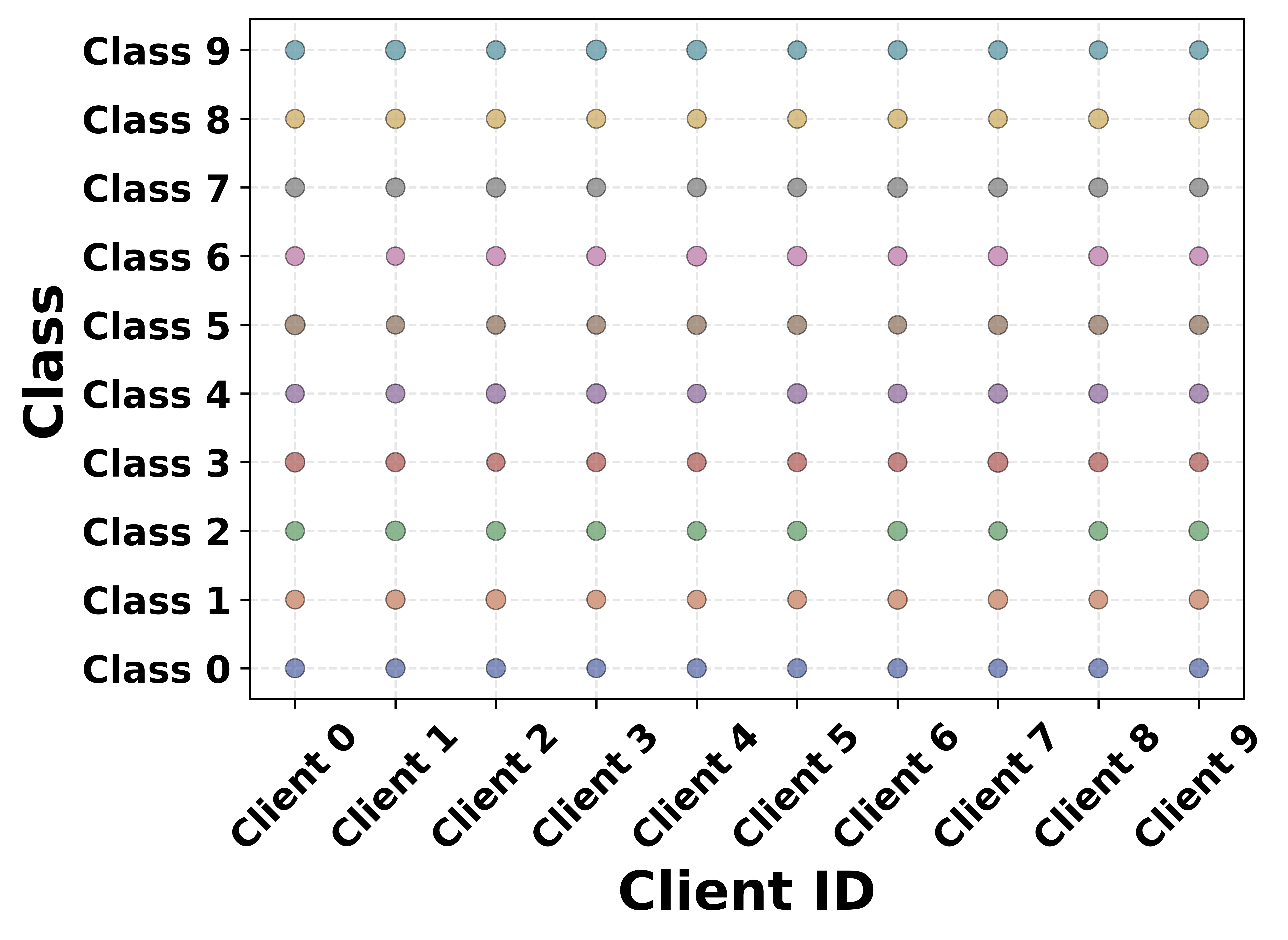}
        \caption{IID}
        \label{fig:image2}
    \end{subfigure}
    
    \par\vspace{0.65cm} 
    
    \begin{subfigure}[b]{0.236\textwidth}
        \centering
        \includegraphics[width=\linewidth, height=4cm]{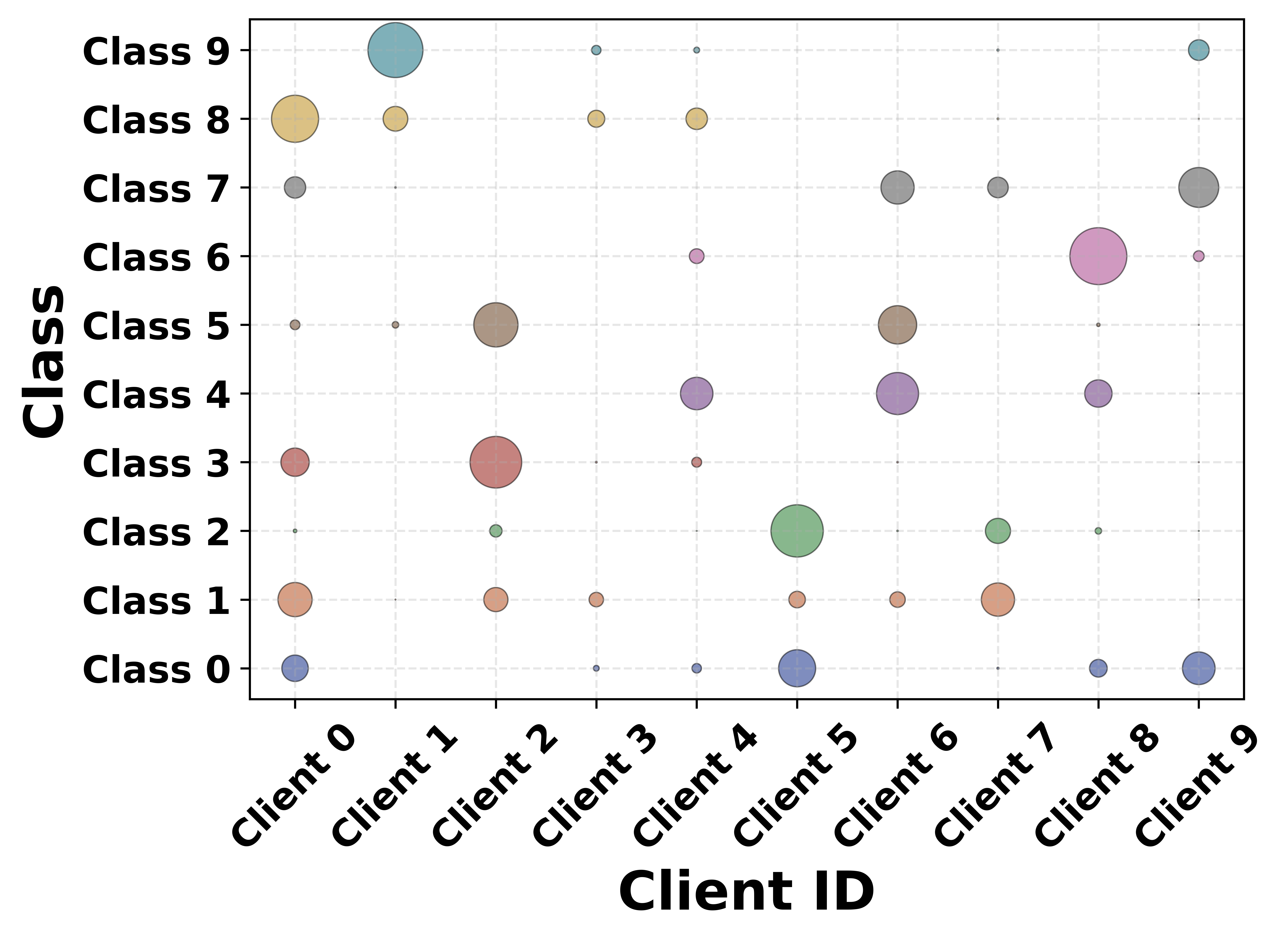}
        \caption{Dir(0.1)}
        \label{fig:image3}
    \end{subfigure}
    \hfill
    \begin{subfigure}[b]{0.236\textwidth}
        \centering
        \includegraphics[width=\linewidth, height=4cm]{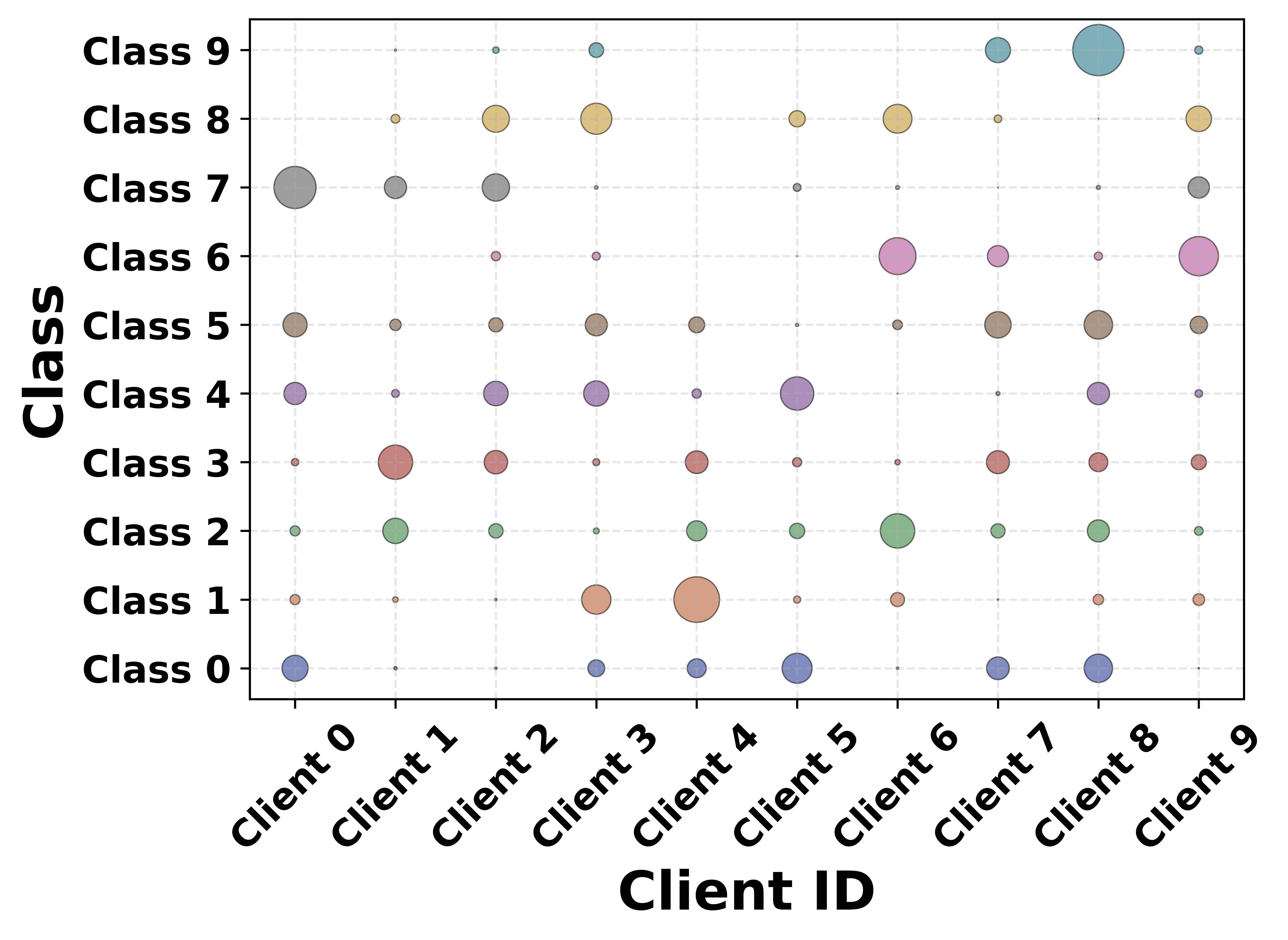}
        \caption{Dir(0.5)}
        \label{fig:image4}
    \end{subfigure}
    
    \caption{Visualization of Data Distributions Across Clients on CIFAR-10 Under Different Heterogeneity Levels}
    \label{fig:data_distribution}
\end{figure}

\textbf{Dataset.} We conduct experiments on three widely used datasets, including Fashion MNIST~\cite{FashionMnist}, CIFAR-10~\cite{Cifar10} and CIFAR-100~\cite{cifar100}.

\begin{table*}[htbp]
    \centering
    \small
    \setlength{\tabcolsep}{3pt}
    \begin{tabular}{c|cccc|cccc|cccc}
    \toprule
     Dataset & \multicolumn{4}{c|}{FASHION MNIST }  & \multicolumn{4}{c|}{CIFAR-10 } & \multicolumn{4}{c}{CIFAR-100} \\

     Setting & IID & Pathological & $Dir(0.5)$ & \textit{Avg} & IID & Pathological & $Dir(0.5)$ & \textit{Avg} & IID & Pathological & $Dir(0.5)$ & \textit{Avg} \\
     
     \midrule
     
    Local & 84.77 & 99.28 & 90.38 & 91.48 & 54.96 & 91.05 & 75.08 & 73.70 & 16.61 & 49.41 & 30.30 & 32.11\\
    
    Fedavg & 89.61 & 96.33 & 88.44 & 91.46 & 67.09 & 67.27 & 63.55 & 65.97 & 31.36 & 24.99 & 31.30 & 29.22\\
    
    FedProx & 89.61 & 94.11 & 88.20 & 90.64 & 67.02 & 58.56 & 63.20 & 62.93 & 30.83 & 25.19 & 29.81 & 28.61 \\ 

    SFL* & 88.91 & 98.81 & 91.15 & 92.96 & 66.36 & 89.43 & 73.85 & \underline{76.55} & 30.82 & 50.23 & 34.87 & \underline{38.64} \\

    CFL & 87.95 & 99.35 & 91.41 & 92.90 & 57.53 & 90.65 & 74.55 & 74.24  & 21.39 & 52.35 & 32.01 & 35.25 \\

    Per-Fedavg & 88.14 & 99.16 & 91.93 & \underline{93.07} & 64.33  & 89.90 & 74.59 & 76.27 & 24.21 & 51.14 & 36.46 & 37.27  \\ 
    
    pFedMe & 84.29 & 98.64 & 88.56 & 90.50 & 47.17 & 81.71 & 66.38 & 65.09 & 12.99 & 33.06 & 23.00 & 23.02 \\
    
    FedAMp & 81.59 & 99.02 & 86.93 & 89.18 & 46.03 & 86.05  & 64.40 & 65.49 & 10.17 & 37.92 & 21.47 & 23.19\\
    
    PFEDHN & 85.30 & 99.05 & 90.38 & 91.58 & 48.3 & 83.69 & 66.12 & 73.02 & 14.54 & 36.00 & 26.46 & 25.67 \\
    
    FedROD & 87.35 & 99.17 & 91.55  & 92.69 & 62.79 & 90.54 & 73.93 & 75.75 & 18.69 & 49.17 & 31.07 & 32.98 \\
    
    KNN-PER & 89.63 & 98.06 & 88.89 & 92.19 & 66.81 & 79.61 & 62.89 & 69.77 & 31.38 & 25.80  & 30.21 & 29.13 \\

    FedAH & 87.99 & 99.34 & 91.88 & \underline{93.07} & 63.29 & 90.15 & 74.39 & 75.94 & 26.53 & 50.81 & 36.81  & 38.05 \\

    \midrule
    pFedGAT(ours) & 89.76 & 99.31 & 91.90 & \textbf{93.66} & 67.17 & 90.73 & 74.80 & \textbf{77.57} & 31.38 &  51.52  &  36.87 & \textbf{39.93}  \\
    \bottomrule
    \end{tabular}
    \caption{Performance comparison of different methods on Fashion MNIST, CIFAR-10, and CIFAR-100 datasets under different data heterogeneity levels. The best-performing results and the second-best results for the \textit{Avg} are \textbf{bolded} and \underline{underlined}, respectively.  }
    \label{tab:main_results}
\end{table*}

\begin{table}[!h]
    \centering
    \small
    \begin{tabular}{cccccc}
        \toprule
        Settings  & $Dir(0.1)$ & $Dir(0.3)$  & $Dir(0.5)$ & \textit{Avg} \\

        \midrule
        Local  & 47.85 & 37.24 & 30.30 & 48.46\\

        Fedavg & 27.93  & 29.80 & 31.30 & 29.68\\

        FedProx & 27.28 & 28.11 & 29.81 & 28.40 \\

        SFL*  & \underline{50.83}  & 40.94  & 34.87 & 42.21 \\

        CFL & 47.78 & 38.28 & 32.01 & 39.36 \\

        Per-Fedavg & 50.15 & \underline{41.61} & 36.46 & 42.74 \\

        pFedMe & 34.55 & 26.65 & 23.00 & 28.07 \\

        FedAMP & 31.47 & 26.79 & 21.47 & 29.58 \\

        PFEDHN & 39.10 & 29.83 & 26.46 & 31.80 \\

        FedROD & 47.58 & 36.46 & 31.07 & 38.37 \\

        KNN-PER & 28.87 & 29.00 & 30.21 & 29.36 \\

        FedAH   & 50.81 & 41.26 & \underline{36.81} & \underline{42.96} \\

        \midrule
        pFedGAT & \textbf{51.32} & \textbf{42.15} & \textbf{36.87} & \textbf{43.45}  \\

        \bottomrule
    \end{tabular}
    \caption{The accuracy (\%) on CIFAR-100 for different data heterogenrity.}
    \label{tab:Dir experiment}
\end{table}

\noindent \textbf{Data heterogeneity.} We investigate several levels of data heterogeneity:
\begin{enumerate}
    \item Homogeneous distribution~\cite{fedavg}: Each data sample is assigned to each client with equal probability $\frac{1}{N}$.

    \item pathological distribution~\cite{fedavg, FedProx, FedAMP}: Each client is assigned data from 2 categories for Fashion MNIST and CIFAR-10 and 20 categories for CIFAR-100.In this setting, the data distributions across clients are highly imbalanced.

    \item Dirichlet distribution~\cite{FedDF, FedMA}: For each class $c$, We sample a probability vector $q_{c} \in \mathbb{R}^{N}$ from a Dirichlet distribution $Dir(\beta)$, where $\beta$ is the distribution parameter. The data of class $c$ is then allocated to client $i$ according to the corresponding element $q_{c,i}$ in the vector $q_{c}$, with allocation done proportionally.

\end{enumerate}
Visualizations of data distributions under varying levels of Non-IID data heterogeneity, including pathological, IID, Dirichlet $\beta=0.1$ and Dirichlet $\beta=0.5$ distributions, are shown in Figure\ref{fig:data_distribution}.

\noindent \textbf{Implementation details} We use a simple CNN network consisting of 3 convolutional layers and 3 fully-connected layers~\cite{MooN}.The optimizer is SGD with learning rate 0.01 and a batch size of 64. For GAT, we use 8 attention heads, with a leaky ReLU slope of 0.2. The GAT optimizer  is also SGD with a learning rate of 0.01.  We consider 50 total communication rounds , with each client running 5 epoches.

\noindent \textbf{Baselines} We compare 12 baselines, including local model training without collaboration (denote as \texttt{Local}), two traditional federated learning methods: \texttt{FedAvg}~\cite{fedavg} and \texttt{FedProx}~\cite{FedProx}, one graph-guided PFL method: \texttt{SFL*}~\cite{sfl} and eight representative PFL methods: \texttt{CFL}~\cite{CFL} clusters clients based on the similarity of their data distributions and performs federated learning within each cluster, \texttt{Per-FedAvg}~\cite{perfedavg} employs meta-learning to enable the global model to adapt to local data distributions with minimal local updates, \texttt{pFedMe}~\cite{pFedMe} imposes an $\ell_{2}$ distance constraint between the global and personalized models, and \texttt{pFedHN}~\cite{pfedhn} trains a hypernetwork on the server to generate personalized model parameters for each client. \texttt{FedRoD}~\cite{FedRoD} optimizes both global and personalized model performance simultaneously, \texttt{KNN-Per}~\cite{KNN-PER} leverages the K nearest client models to optimize the local model, and \texttt{FedAH}~\cite{FEDAH} optimizes the aggregation process for both the feature extraction layer and the classification layer.

\begin{table}[!h]
    \centering
    \small
    \begin{tabular}{cccc}
        \toprule
        Client Number  & 10 & 25  & 50  \\

        \midrule
        Local  & 83.89 & 88.97 & 83.98 \\

        Fedavg & 61.11 & 64.41 & 58.58   \\

        FedProx & 61.64 & 60.69 & 58.53  \\
        
        SFL* & 83.26 & 88.58 & 82.47  \\

        CFL & 83.55 & 89.51 & 84.68 \\

        Per-Fedavg & \underline{84.01} & \underline{89.82} & 84.02\\

        pFedMe & 75.64 & 87.29 & 82.23 \\

        FedAMP & 75.58 & 86.72 & 81.64 \\

        PFEDHN & 77.22 & 80.01 & 74.74  \\

        FedROD & 83.30 & 89.61 & 84.73  \\

        KNN-PER & 72.06 & 84.77 & 78.84 \\

        FedAH & 83.23 & 89.39 & \underline{84.80} \\

        \midrule
        pFedGAT & \textbf{84.23} &  \textbf{89.91}   &   \textbf{85.05} \\

        \bottomrule
    \end{tabular}
    \caption{The accuracy (\%) on CIFAR-10 for different client numbers.}
    \label{tab:Clients experiments}
\end{table}

\subsection{Main Experiment}
\label{sec:main experiment}
To demonstrate the superiority of our proposed \texttt{pFedGAT} algorithm, we compared it against 12 baseline methods on the three datasets, under three different data distributions. The detailed results are shown in Table \ref{tab:main_results}.

From the table, we make the following conclusions:

\begin{itemize}
    \item[i)] Across all three datasets and under all data distributions, our \texttt{pFedGAT} algorithm consistently outperforms or matches the performance of other methods. In contrast, other methods show strong performance only in specific scenarios, highlighting the robustness of our approaches in diverse settings.

    \item[ii)] Since real-world data distributions are unknown, we compare the average performance across various scenarios with that of other methods~\cite{pFedGraph}. Our algorithm consistently achieves the best overall performance, demonstrating their effectiveness across different conditions.

    \item[iii)] The results from the table indicate that \texttt{SFL*} is a robust baseline, likely due to its consideration of the graph structure formed among client models, which enables more effective regulation of the parameter aggregation process.

\end{itemize}

\subsection{Different Degrees of Heterogeneity}
To demonstrate that our method performs well under various degrees of Non-IID scenarios, we modified the parameter $\beta$ in $Dir(\beta)$ on the CIFAR-100 dataset to control the data heterogeneity among clients. We then tested various methods and the results are shown in Table ~\ref{tab:Dir experiment}. It is clear that \texttt{pFedGAT} consistently performs well across different levels of data heterogeneity.



\subsection{Different Number of Clients}
To demonstrate that our method can still capture graph relationships among clients as the number of clients increases, we varied the number of clients, set the data distribution to $Dir(0.1)$, and conducted experiments on the CIFAR-10 dataset using different methods. The results are presented in Table~\ref{tab:Clients experiments}. As shown, our method consistently captures inter-client graph relationships even as the number of clients grows, maintaining strong performance.

From the table, we observe a phenomenon: when the number of clients increases from 10 to 25, the accuracy rises; however, when it further increases from 25 to 50, the accuracy declines. In federated learning, a larger number of data samples on a client generally leads to more thorough training and better performance; conversely, fewer label categories result in lower task difficulty and improved performance. Given that our data distribution is set to 
$Dir(0.1)$, which exhibits significant Non-IID characteristics, it is evident that when the number of clients is too small, each client has an excessive number of label categories, increasing task difficulty. Conversely, when the number of clients is too large, the data samples per client are insufficient, leading to inadequate training. When the number of clients is 25, an optimal balance between the number of data samples and label categories is achieved, resulting in the best performance.

\vspace{0.1cm}
\section{Analysis}
\label{sec:analysis}

\begin{figure}[!htbp]
    \centering
    \begin{subfigure}[b]{0.22\textwidth}
        \includegraphics[width=\linewidth]{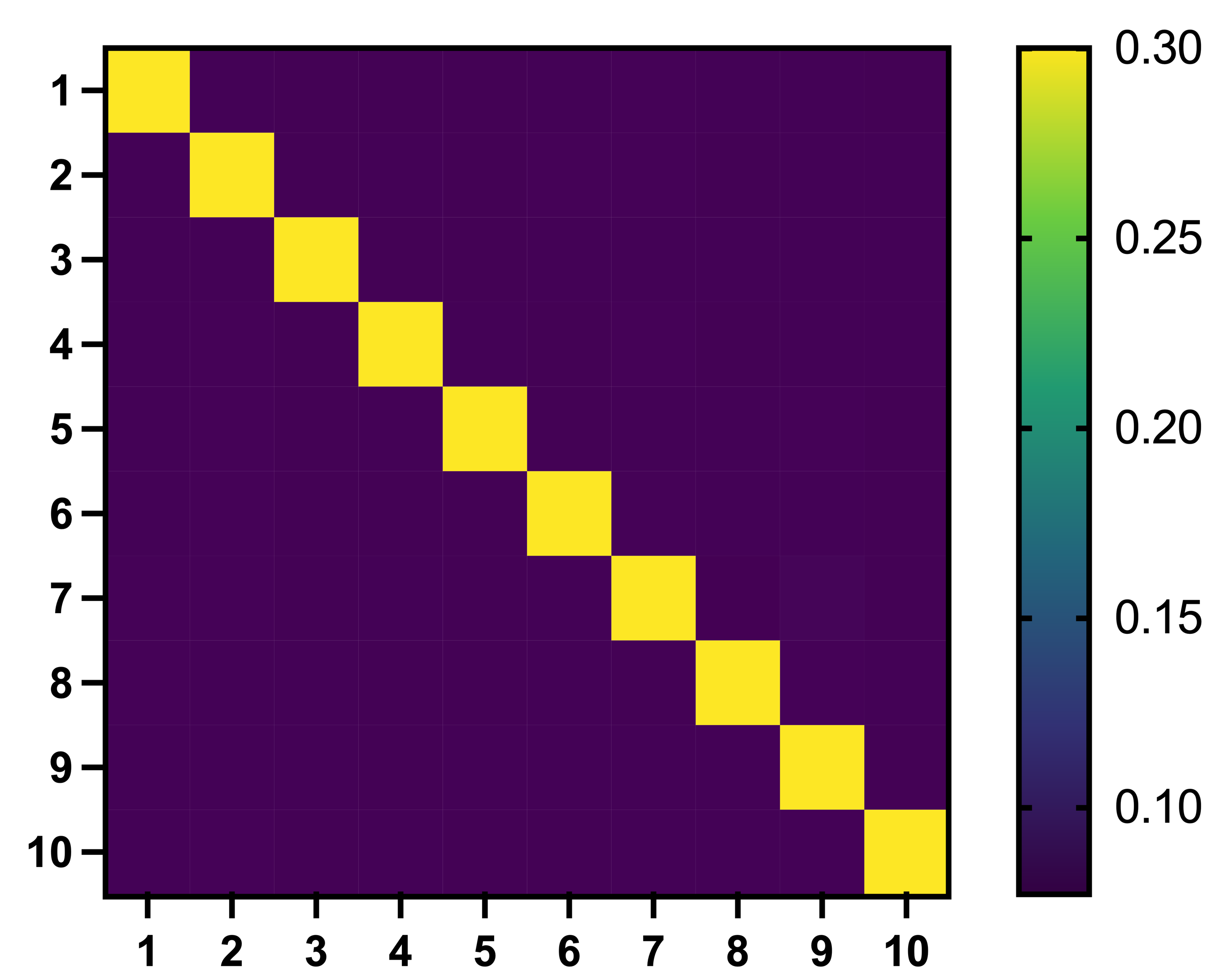}
        \caption{FedAMP(IID)}
        \label{fig:IID_FedAMP}
    \end{subfigure}
    \hfill
    \begin{subfigure}[b]{0.22\textwidth}
        \includegraphics[width=\linewidth]{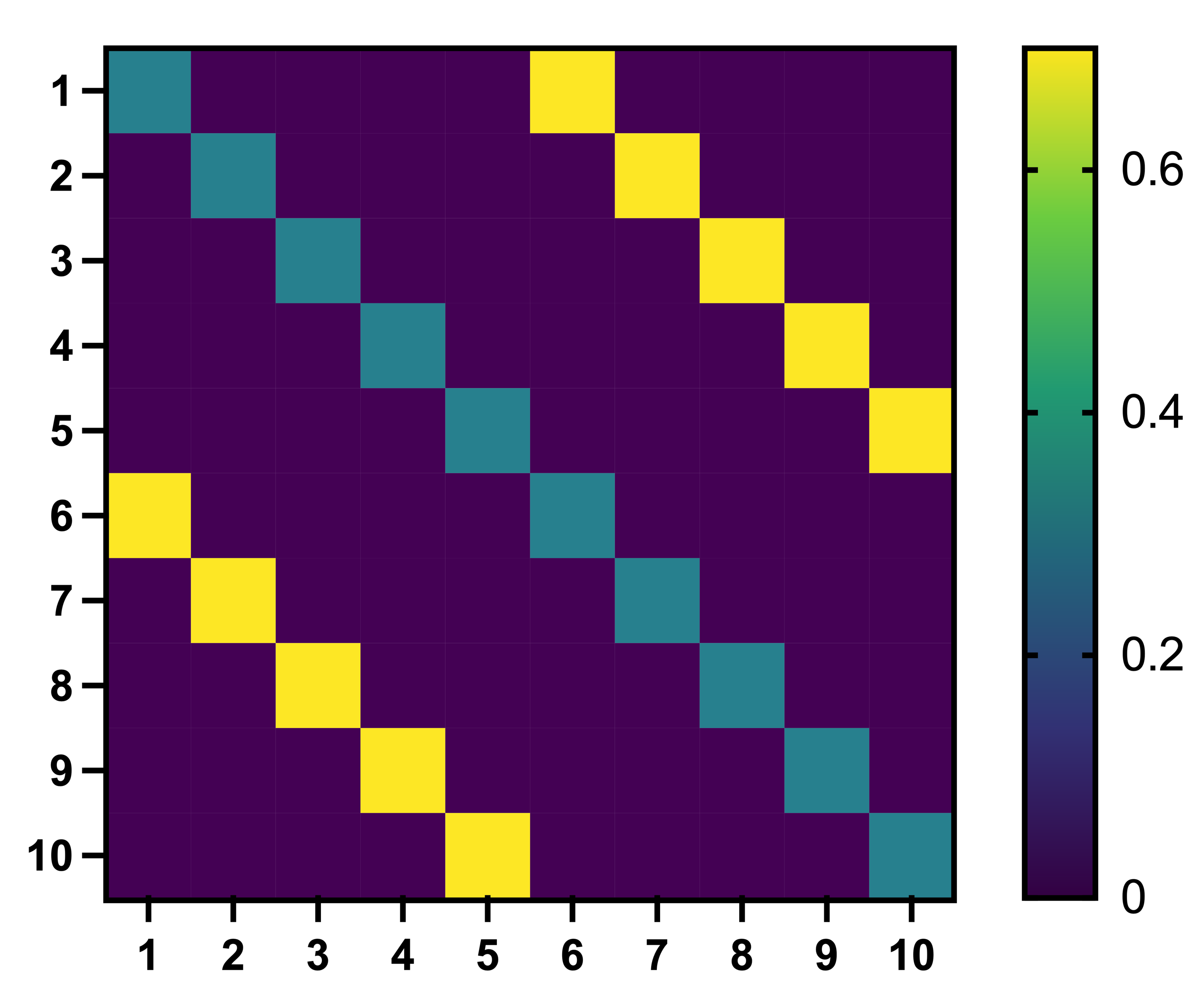}
        \caption{FedAMP(Path)}
        \label{fig:Path_FedAMP}
    \end{subfigure}

    \vspace{0.65cm}
    
    \begin{subfigure}[b]{0.22\textwidth}
        \includegraphics[width=\linewidth]{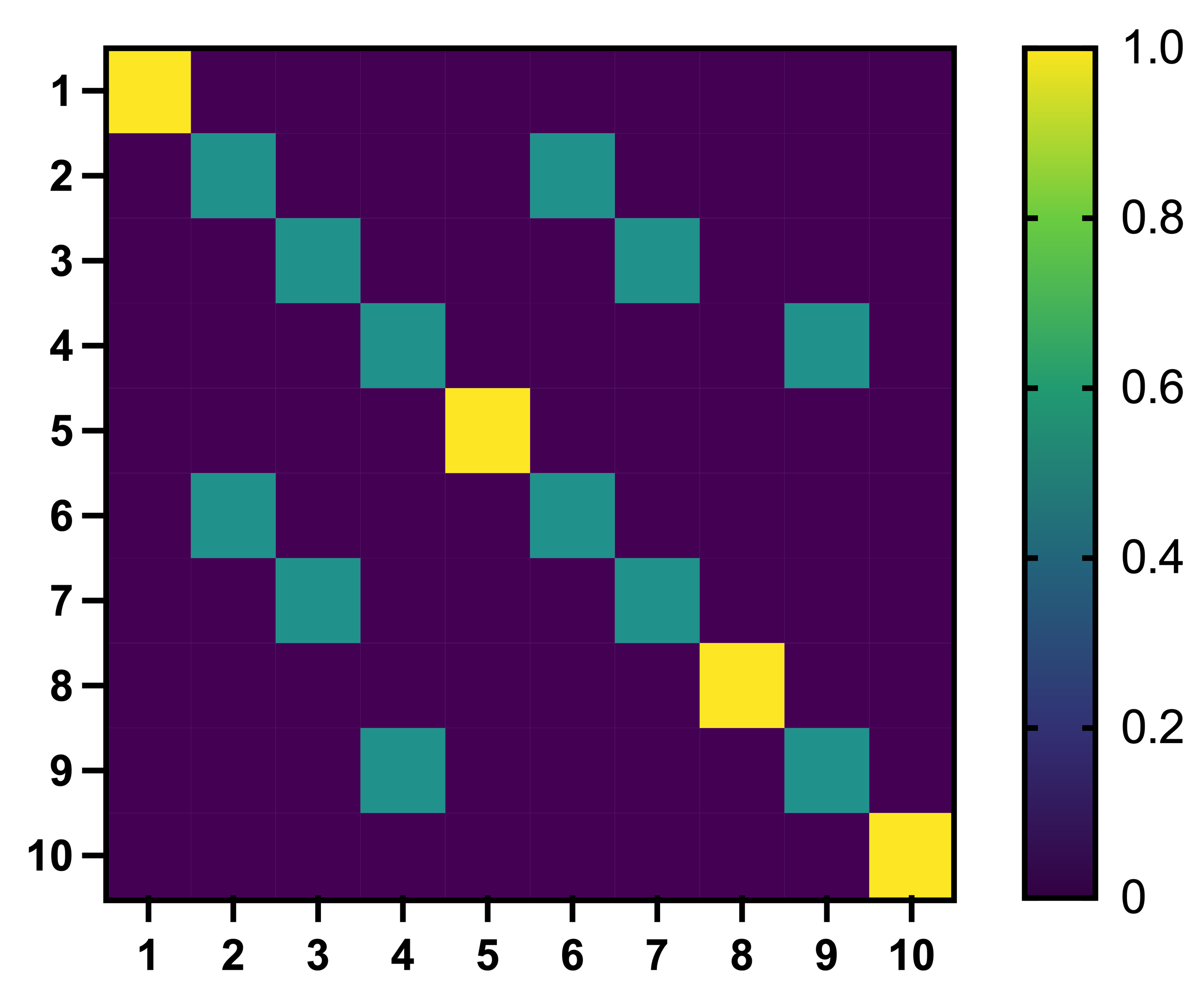}
        \caption{CFL(IID)}
        \label{fig:IID_CFL}
    \end{subfigure}
    \hfill
    \begin{subfigure}[b]{0.22\textwidth}
        \includegraphics[width=\linewidth]{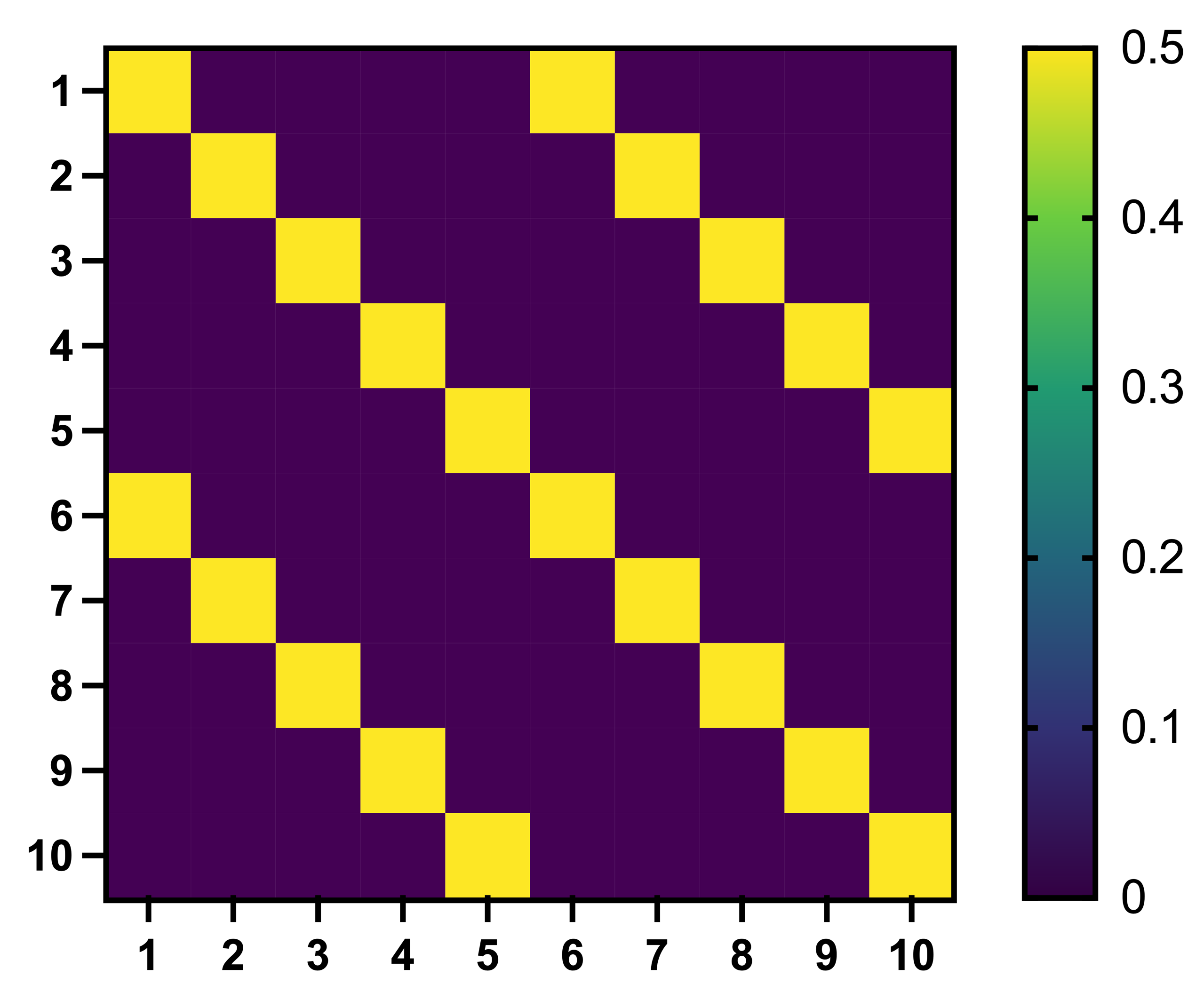}
        \caption{CFL(Path)}
        \label{fig:Path_CFL}
    \end{subfigure}

    \vspace{0.65cm}
    
    \begin{subfigure}[b]{0.225\textwidth}
        \includegraphics[width=\linewidth]{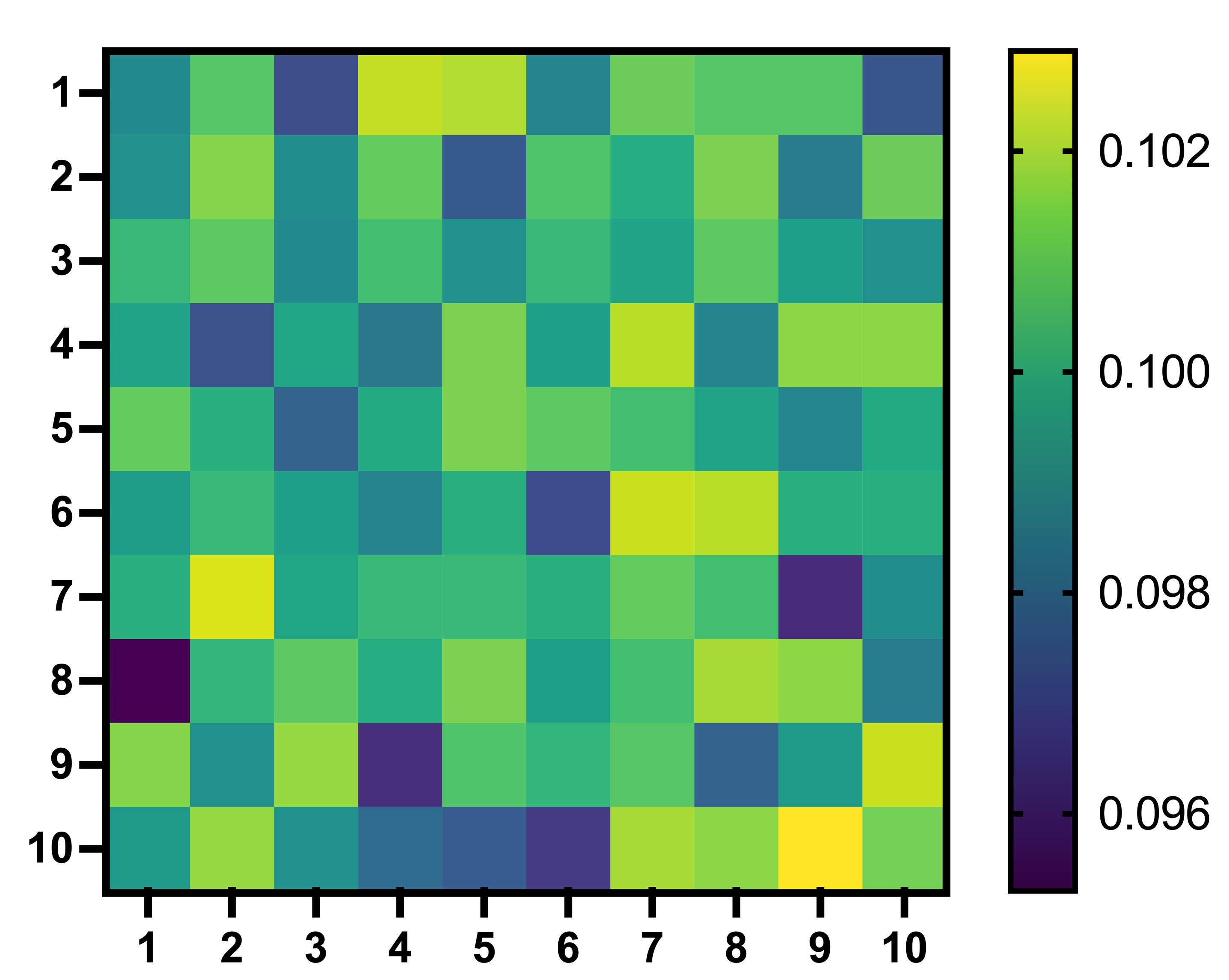}
        \caption{pFedGAT(IID)}
        \label{fig:IID_pFedGAT}
    \end{subfigure}
    \hfill
    \begin{subfigure}[b]{0.22\textwidth}
        \includegraphics[width=\linewidth]{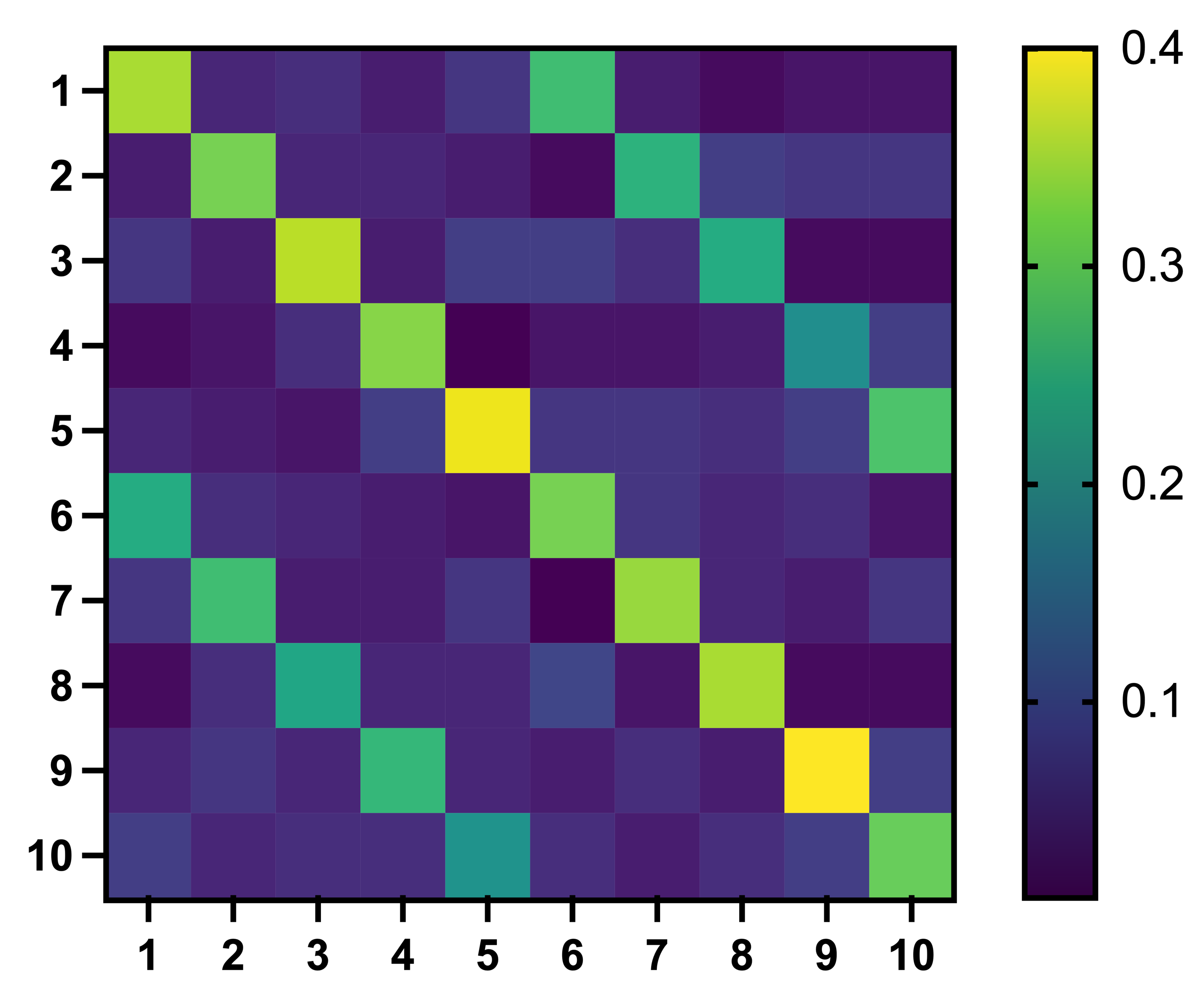}
        \caption{pFedGAT(Path)}
        \label{fig:Path_CFL}
    \end{subfigure}
    
    \caption{Visualization of the weight allocation matrices under IID and Pathological data distribution.}
    \label{fig:heatmap}
    \vspace{0.2cm}
\end{figure}

\begin{figure}[!h] 
    \centering
    \begin{subfigure}[b]{0.23\textwidth} 
        \includegraphics[width=\linewidth]{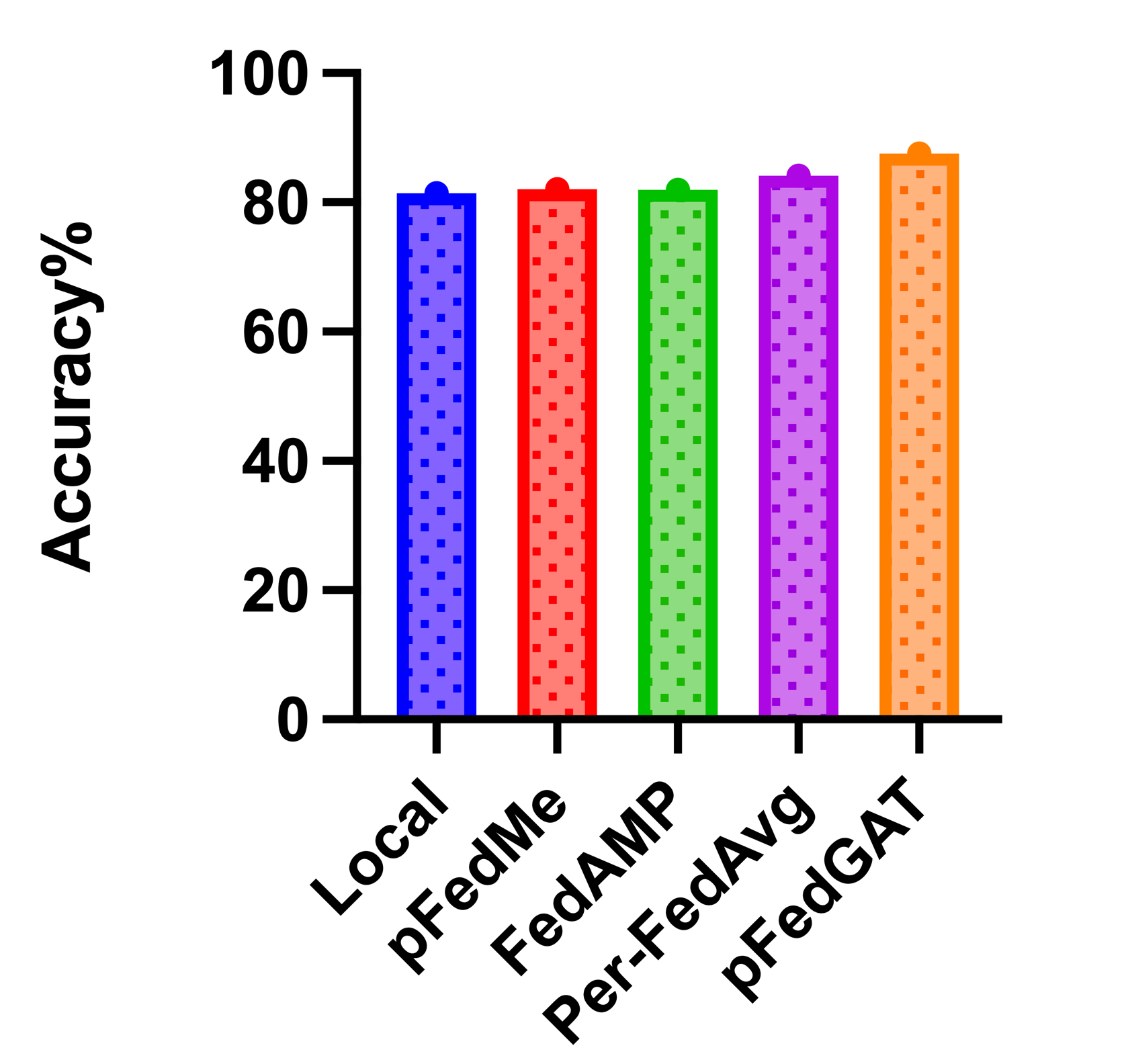}
        \caption{CIFAR-10}
    \end{subfigure}
    \hfill
    \begin{subfigure}[b]{0.2\textwidth}
        \includegraphics[width=\linewidth]{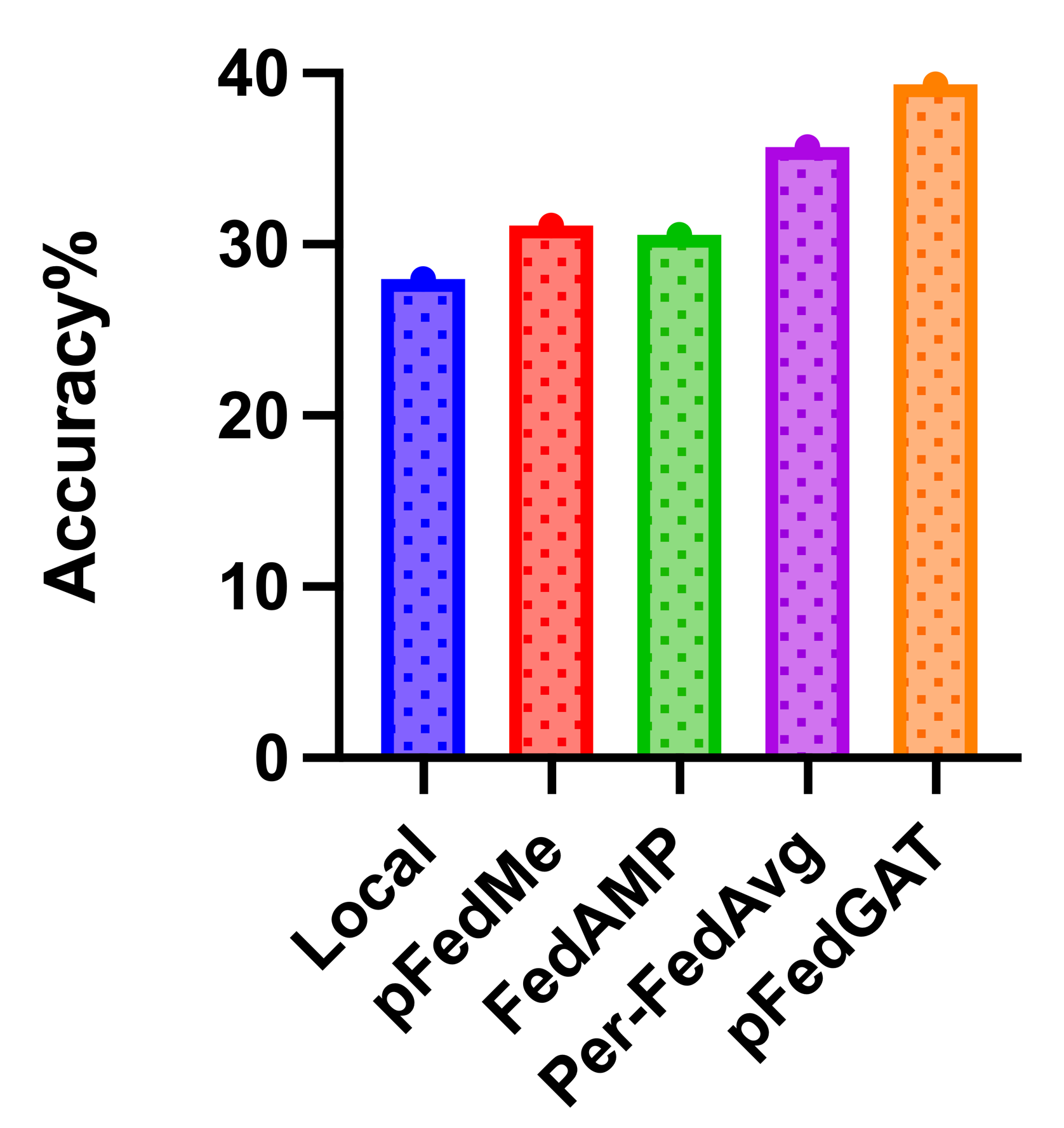}
        \caption{CIFAR-100}
    \end{subfigure}

    \vspace{0.65cm}

    \caption{A Comparison of Generalization Performance on CIFAR-10 and CIFAR-100. Our method consistently achieves the highest performance across all scenarios.}
    \label{fig:generalize}
\end{figure}

Based on the experiments above, we further discuss the following three questions:

\begin{itemize}
    \item Q1: Can the graph structure enhance the optimization of collaborative relationships among clients? (Section \ref{sec:Graph Structure}, Table \ref{tab:without graph})

    \item Q2: Can GAT capture the collaborative relationships among clients? (Section \ref{sec:Visualization}, Fig \ref{fig:heatmap})  

    \item Q3: Can our method generalize to new clients? (Section \ref{sec:Generalizability}, Fig \ref{fig:generalize})
\end{itemize}

\subsection{Effectiveness of the Graph Structure}
\label{sec:Graph Structure}

In our proposed method, we employ a GAT to model the graph structure emerging from the client models, dynamically generating a weight allocation matrix. The GAT parameters are subsequently optimized according to Equation~\ref{eq:optimize}, ensuring that the resulting matrix more accurately reflects the collaborative relationships among clients. However, a natural question arises: why not directly optimize the weight allocation matrix to minimize $\mathcal{L}$? To address this intuitively, we conducted a comparative experiment, with the results detailed in Table~\ref{tab:without graph}.

The experimental outcomes reveal that our GAT-based approach consistently surpasses the performance of directly optimizing the weight allocation matrix. This superiority can be attributed to the following key factors: GAT optimizes its parameters to indirectly refine the weight matrix through nonlinear transformations and multi-head attention mechanisms, enabling exploration of a broader and more expressive solution space. In contrast, direct optimization of the matrix via gradient descent typically involves linear adjustments, which lack the capacity to capture complex, high-order relationships among clients. Furthermore, GAT leverages graph structures and attention mechanisms to effectively uncover nonlinear dependencies, such as community structures or latent patterns, leading to a more discriminative and adaptive weight allocation. This dynamic learning capability enhances our method’s adaptability and performance, particularly in Non-IID settings.



\subsection{Visualization of Weight Allocation Matrices}
\label{sec:Visualization}
In this section, we utilize the CIFAR-10 dataset with data distributions set to IID and Pathological, and a client number of 10, to visualize the weight allocation matrices computed by \texttt{FedAMP}, \texttt{CFL}, and \texttt{pFedGAT}, with the results presented in Figure \ref{fig:heatmap}.

The visualization reveals the following insights: i) Under the IID data distribution, \texttt{FedAMP}'s attention mechanism fails to capture the uniformity of data distributions across clients, resulting in each client predominantly collaborating with itself. Similarly, \texttt{CFL} does not facilitate equitable collaboration among all clients, instead erroneously grouping them into several clusters for intra-cluster cooperation. In contrast, the weight allocation matrix computed by \texttt{pFedGAT} exhibits values consistently around 0.1, demonstrating that our method accurately recognizes the consistent data distributions across clients. ii) Under the Pathological data distribution, all three methods assign greater collaboration weights to clients with similar data distributions. However, \texttt{FedAMP} and \texttt{CFL} overlook collaboration with clients having dissimilar data distributions. \texttt{pFedGAT}, on the other hand, assigns larger weights to clients with similar data distributions and smaller weights to those with dissimilar distributions, enabling our method to avoid overfitting to local data while reducing the learning of detrimental knowledge. These observations indicate that \texttt{pFedGAT} effectively captures the collaborative relationships among clients.

\begin{table}[t]
    \centering
    \small
    \setlength{\tabcolsep}{2pt}
    \begin{tabular}{lccccc}
    \toprule
    Settings & IID & Pathological & $Dir(0.1)$ & $Dir(0.3)$ & $Dir(0.5)$    \\
    \midrule
    without GAT & 30.23 & 50.75  & 49.87  & 41.15 & 36.36  \\
    pFedGAT & 31.38 & 51.52  &  51.32 & 42.15 & 36.87  \\
    \bottomrule
    \end{tabular}
    \caption{A comparison between our method and the approach of directly optimizing the weight allocation matrix. The experiment was conducted on the CIFAR-100 dataset with five different data distribution scenarios. }
    \label{tab:without graph}
\end{table}

\subsection{Generalizability to New Clients}
\label{sec:Generalizability}
In traditional Graph Attention Network (GAT) models, GAT efficiently captures the importance relationships between newly added graph nodes and existing nodes. In real-world federated learning scenarios, it is common for new clients to join the process during training, necessitating rapid adaptation to their local data distributions. To evaluate whether our GAT model can similarly and effectively identify collaborative relationships between these new clients and existing ones within a federated learning framework, we conducted a targeted experiment to assess its generalization capability. We compared our approach with several baseline methods through a simulated scenario. Specifically, we trained on the CIFAR-10 and CIFAR-100 datasets with a data distribution set to $Dir(0.1)$. In the initial 40 rounds, only 8 clients participated in federated learning. In the subsequent 3 rounds,  2 new clients were integrated into the federated learning process. We compare \texttt{pFedGAT} with \texttt{Local} training, \texttt{pFedMe}, \texttt{FedAMP}, and also \texttt{Per-FedAvg}, the latter of which aims to train an adaptable model that can quickly adjust to the local data of each client. The result can be seen in Fig \ref{fig:generalize}.

Given that training continued for only three rounds after the introduction of new clients, the final personalized accuracy of these newly joined clients serves as a reliable indicator of each method's ability to generalize to new clients. The visualizations reveal that the \texttt{Local} training method performs poorly, likely due to its reliance on isolated local training without inter-client collaboration, compounded by the insufficient size of local datasets to meet training demands. Among the federated learning algorithms, \texttt{pFedGAT} achieves best performance, suggesting that our method effectively captures the collaborative relationships between new clients and existing ones. Notably, \texttt{pFedGAT} demonstrates strong generalization to new clients, highlighting its adaptability in this context.

\section{Conclusions}

In personalized federated learning, a key challenge is regulating the model aggregation process. This paper pioneers the use of Graph Attention Networks (GAT) to dynamically adjust aggregation weights based on client importance, achieving effective personalization. We also introduce an end-to-end optimization method where clients test the aggregated model locally, backpropagating the loss to the server, which then refines the GAT model for precise weight adjustments. This optimization approach enables the server to receive real-time feedback from clients without increasing communication overhead. Extensive experiments across diverse data distributions and datasets validate the superiority of our \texttt{pFedGAT}.

\vspace{0.1cm}
{
    \small
    \bibliographystyle{ieeenat_fullname}
    \bibliography{main}
}

\end{document}